\documentclass[10pt,twocolumn,letterpaper]{article}

\usepackage[pagenumbers]{cvpr}      

\usepackage{graphicx}
\usepackage{amsmath}
\usepackage{amssymb}
\usepackage{booktabs}
\usepackage{balance}
\usepackage{multirow}
\usepackage{multicol}
\usepackage{colortbl}
\definecolor{mygray}{gray}{.9}
\usepackage{mwe}
\usepackage{algorithmic,float}
\usepackage[linesnumbered,ruled,vlined,noline]{algorithm2e}
\usepackage{etoolbox}
\usepackage{relsize}
\usepackage{comment}
\usepackage{bm}
\usepackage{enumitem}
\usepackage{adjustbox}
\usepackage{soul}

\usepackage[pagebackref,breaklinks,colorlinks]{hyperref}

\usepackage[capitalize]{cleveref}
\crefname{section}{Sec.}{Secs.}
\Crefname{section}{Section}{Sections}
\Crefname{table}{Table}{Tables}
\crefname{table}{Tab.}{Tabs.}

\def\cvprPaperID{3902} 
\def\confName{CVPR}
\def\confYear{2023}

\begin{document}

\title{Binarizing Sparse Convolutional Networks for Efficient Point Cloud Analysis}

\author{Xiuwei Xu\textsuperscript{1,2}, ~Ziwei Wang\textsuperscript{1,2}, ~Jie Zhou\textsuperscript{1,2}, ~Jiwen Lu\textsuperscript{1,2}\thanks{Corresponding author.}\\
\textsuperscript{1}Department of Automation, Tsinghua University, China \\
~\textsuperscript{2}Beijing National Research Center for Information Science and Technology, China \\
{\tt\small \{xxw21, wang-zw18\}@mails.tsinghua.edu.cn;}
{\tt\small \{jzhou, lujiwen\}@tsinghua.edu.cn} \\
}
\maketitle

\begin{abstract}
In this paper, we propose binary sparse convolutional networks called BSC-Net for efficient point cloud analysis. We empirically observe that sparse convolution operation causes larger quantization errors than standard convolution. 
However, conventional network quantization methods directly binarize the weights and activations in sparse convolution, resulting in performance drop due to the significant quantization loss.
On the contrary, we search the optimal subset of convolution operation that activates the sparse convolution at various locations for quantization error alleviation, and the  performance gap between real-valued and binary sparse convolutional networks is closed without complexity overhead.
Specifically, we first present the shifted sparse convolution that fuses the information in the receptive field for the active sites that match the pre-defined positions. Then we employ the differentiable search strategies to discover the optimal opsitions for active site matching in the shifted sparse convolution, and the quantization errors are significantly alleviated for efficient point cloud analysis.
For fair evaluation of the proposed method, we empirically select the recently advances that are beneficial for sparse convolution network binarization to construct a strong baseline. 
The experimental results on ScanNet and NYU Depth v2 show that our BSC-Net achieves significant improvement upon our srtong baseline and outperforms the state-of-the-art network binarization methods by a remarkable margin without additional computation overhead for binarizing sparse convolutional networks.
\end{abstract}
  
  
\section{Introduction}
3D deep learning on point clouds~\cite{qi2017pointnet,qi2017pointnet++,graham20183d,choy20194d} has been widely adopted in a wide variety of downstream applications including autonomous driving, AR/VR and robotics due to its strong discriminative power and generalization ability. In these applications, real-time interaction and fast response are required to guarantee safety and practicality. 

\begin{figure}[t]
    \centering
    \includegraphics[width=1.0\linewidth]{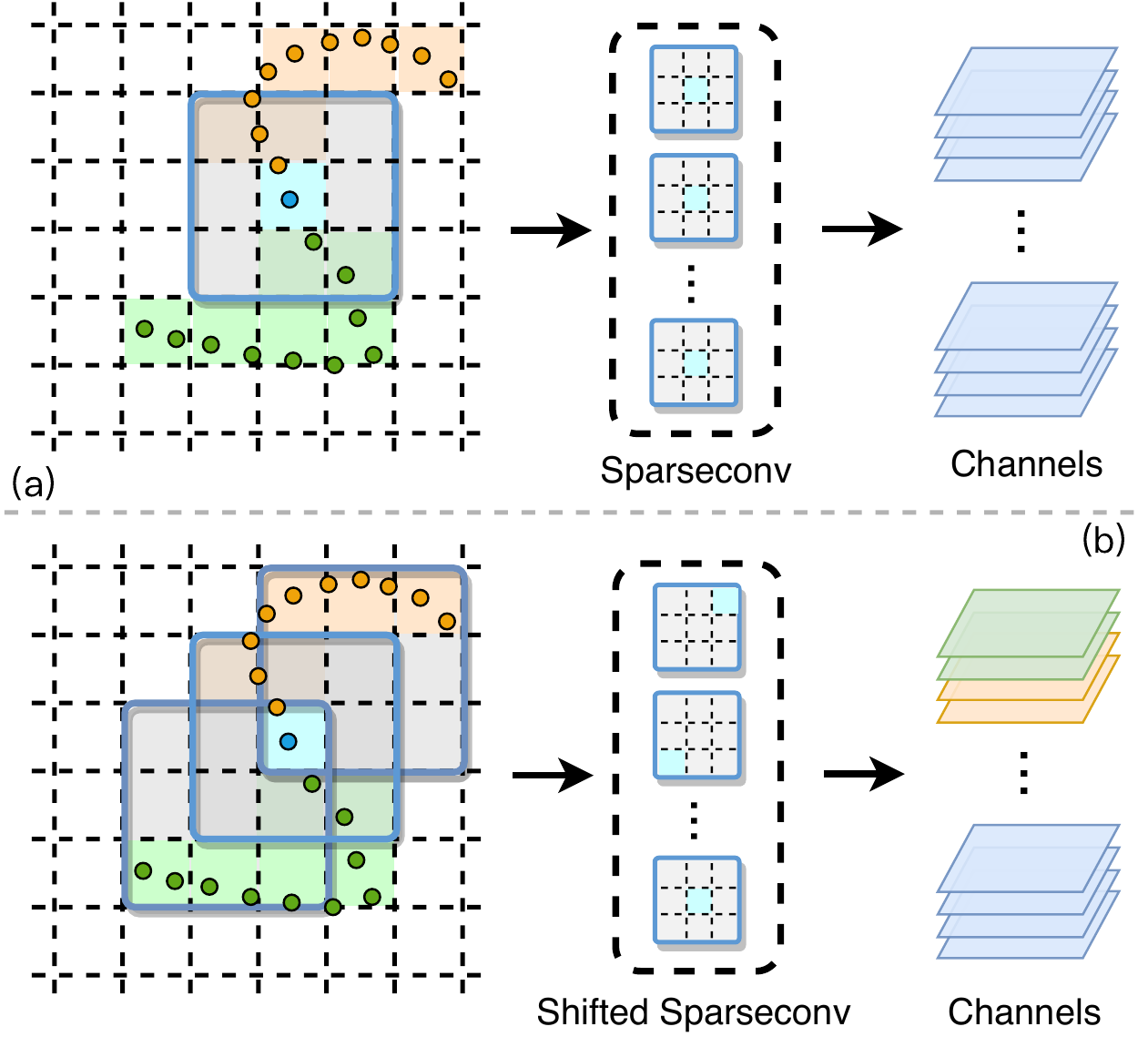}
    \vspace{-4mm}
    \caption{Demonstration of sparse convolution and the proposed shifted sparse convolution. (a) Sparse convolution only operates when the center of kernel slides over the active sites. (b) Our shifted sparse convolution performs different operations for each group of output channels, which brings more information from the neighbor active sites.}
    \vspace{-2mm}
    \label{fig:SSSC}
\end{figure}

Submanifold sparse convolution (we call it "sparse convolution" for short in the rest of this paper)~\cite{graham20183d} is one of the most popular and basic operator for point cloud analysis, which first voxelizes the point clouds and then applies 3D convolution on the voxels while keeping the same sparsity pattern throughout the layers of the network.
Sparse convolution is widely adopted in most state-of-the-art architectures for point cloud analysis and so it is desirable to further improve its efficiency for more practical application. 
We opt for architecture-agnostic methods such as employing network binarization to achieve this goal. 
Binarized neural networks~\cite{liu2020react,wang2021learning} restrict the bitwidth of weights and activations to only one bit and substitute the multiplication-addition by xnor-bitcount operations, which decreases the storage and computational cost by $32\times$ and $64\times$ respectively.
We empirically find sparse convolution operation brings larger quantization errors compared to standard convolution, which leads to significant performance degradation when directly applying existing network binarization methods due to the large quantization errors.

In this paper, we present BSC-Net to learn binary sparse convolutional networks for efficient point cloud analysis in resource-exhaustive scenarios. Instead of directly binarizing the weights and activations in sparse convolutional networks, we search the optimal subset of convolution operation that activates the sparse convolution at various locations for binarization.
The acquired convolution patterns significantly reduces the quantization errors in deployment, and achieves remarkable performance enhancement without extra computational cost. 
More specifically, we propose the shifted sparse convolutional networks whose convolution operations are activated for active sites consistent with the pre-defied locations, and the optimal positions for active site matching across various channels are obtained via differentiable search strategies. Therefore, the quantization errors in the fixed convoltion operations are significantly alleviated by leveraging the shifted sparse convolution with the searched active site matching locations. 
Moreover, we empirically select the recently advances that are beneficial for sparse convolution network binarization to construct a strong baseline.
Extensive experimental results on ScanNet and NYU Depth v2 for semantic segmentation of point clouds show that our BSC-Net reduces the operations per second (OPs) by $92.4\%$ with only $3\%$ mIOU degradation.


\section{Related Work}
\textbf{Network quantization:} Network quantiztaion has been widely studied in computer vision due to the significant enhancement in storage and computation efficiency. Conventional methods can be divided into two categories including networks in one bit and multiple bits. For the first regard, weights and activations in networks are binarized with extremely high compression ratio. Hubara \emph{et al.}~\cite{hubara2016binarized} and Courbariaux \emph{et al.}~\cite{courbariaux2016binarized} substituted the add-multiplication (MAC) of real-valued models by xnor-bitcount operations, and emplyed the straight-through estimators (STE) to update the network parameters with back-propagation. Rastegari \emph{et al.}~\cite{rastegari2016xnor} further presented scaling factors for weights and activations quantization error minimization. To recover the capacity degradation caused by aggressive quantization, Liu \emph{et al.}~\cite{liu2018bi} added extra shortcut connections between consecutive layers to diversify the feature maps. Bulat \emph{et al.}~\cite{bulat2020bats} modified the search space and strategy with stability regularization for the optimal architecture acquisition of binary networks. Qin \emph{et al.}~\cite{qin2020bipointnet} maximized the information entropy in binary features and leveraged learnable scaling factors for information retention in point cloud analysis. Since increasing the weight and activation bitwidths can significantly enhance the model capability, networks in multiple bits are proposed for better performance. Choi \emph{et al.}~\cite{choi2018pact} optimized the activation clipping threshold to find the right quantization scale. Zhang \emph{et al.}~\cite{zhang2018lq} further searches the optimal quantizer basis and encoding for accurate quantization. Lee \emph{et al.}~\cite{lee2021network} adaptively scaled the gradient element in STE to calibrate the direction of parameter update with minimal discretization errors. However, multi-bit networks still suffers from heavy computational and storage cost. Directly applying existing network binarization methods to submanifold sparse convolution destructs the geometric structure in the scene and degrades the feature informativeness significantly.

\noindent\textbf{Sparse convolution networks:} Increased attention has been paid to 3D deep learning on point cloud in recent years, which is important for autonomous driving, AR/VR and robotics. Due to the unordered property of point cloud data, voxelizing the points and applying convolution on 3D grids is a natural solution~\cite{chang2015shapenet,qi2016volumetric,zhou2018voxelnet}. However, as point cloud only covers the surfaces of objects/scenes and the most space in 3D scans is empty, the dense volumetric representation is inherently inefficient. Moreover, the computational cost and memory requirement both increase cubically with voxel resolution, thus making it infeasible to train a voxel-based model with high-resolution inputs. To handle this problem, sparse convolution~\cite{graham2014spatially,engelcke2017vote3deep} were proposed to restricts computation and storage to “active” sites (i.e.\ voxels which are not empty). However, as convolutional operator will increase the number of active sites with each layer, the feature sparsity is reduced accordingly. To further improve the efficiency of sparse CNNs, Graham \emph{et al.}~\cite{graham20183d} introduced submanifold sparse convolution, which only conducts convolution when the center of kernel slides over active sites and keeps the same level of sparsity throughout the network. This made it practical to train networks with more convolution layers, such as UNet~\cite{ronneberger2015u}, FCN~\cite{long2015fully} and ResNets~\cite{he2016deep}. Choy \emph{et al.}~\cite{choy20194d} proposed Minkowski Engine, which extended submanifold sparse convolution to higher dimensions. 
Tang \emph{et al.}~\cite{tang2020searching} further combined point-based model and sparse CNN to achieve both accurate and efficient 3D perception for large scale scenes. In particular, sparse convolutional networks are able to adopt common deep architectures from 2D vision, which can help standardize deep learning for point cloud, and they are widely utilized in state-of-the-art models for various tasks, such as semantic segmentation~\cite{nekrasov2021mix3d}, object detection~\cite{rukhovich2021fcaf3d,wang2022cagroup3d} and instance segmentation~\cite{liang2021instance,schult2022mask3d,vu2022softgroup}.

\noindent\textbf{Differentiable search:} In order to reduce the search complexity during the exploration process, differentiable search has been widely used in network architecture search~\cite{liu2018darts}, mixed-precision quantization~\cite{cai2020rethinking,wang2022quantformer} and continual learning~\cite{zhang2022continual}. During differentiable search, the superstructure containing all choices as different components is constructed, where the importance weights for each branch is optimized with gradient descent for optimal solution acquisition. Liu \emph{et al.}~\cite{liu2018darts} relaxed the space of network architectures to be continuous, and jointly optimized the branch importance weights and parameters of the hypernet for network architecture search. Cai \emph{et al.}~\cite{cai2020rethinking} assigned different bitwidths to various branches in the supernet for mixed-precision quantization, and chose the bitwidth in the component with the largest importance weight to be the quantization strategy during inference to achieve the optimal accuracy-complexity trade-off. Guan \emph{et al.}~\cite{guan2020differentiable} updated the feature weights through the presented bridge loss which enhanced the knowledge distillation between the students and teachers. In this paper, we extend differentiable search for the discovery of optimal position for active site matching in shifted sparse convolution, where the search cost is significantly reduced for exploration in large space.
  
\section{Approach}
In this section, we first briefly introduce the preliminary concept of sparse convolution and network binarization. Then we conduct experiments to show the quantization errors of network binarization methods in different convolution patterns, and introduce the shifted sparse convolution (SFSC) operation which is activated for sites in various locations of the receptive field. Finally, we demonstrate the differentiable search to discover the optimal position for active site matching in SFSC, and construct the BSC-Net with alleviated quantization errors and enhanced performance.


\subsection{Preliminaries}\label{a1}
Let $\bf x_u$ be an input feature vector of an active site, located at $3$-dimensional coordinates ${\bf u} \in \mathbb{R}^{D}$. 
As shown in Figure \ref{fig:SSSC}(a), the general sparse convolution~\cite{graham20183d,choy20194d} $F_0$ by a kernel for $\bf x_u$ is formulated as: 
\vspace{-2mm}
\begin{equation}
    F_0(\boldsymbol W,{\bf x_u})=\sum_{{\bf i}\in N^D({\bf u})}{\boldsymbol W_{\bf i}{\bf x_{u+i}}}
    \vspace{-2mm}
\end{equation}
where $N^D({\bf u})$ denotes the list of offsets in the 3-dimensional cube centered at origin $\bf u$. 
The convolution kernel can be break down and assigned to each offset parameterized by $\bf W_i$. 

Sparse convolution is a practical substitution for vanilla 3D convolution, and skips the non-active regions that only operates when the center of convolutional kernel covers active voxels. Specifically, active voxels are stored as sparse tensors for the fixed convolution operations, where all active synapes between input and output voxels are found to perform convolution. Therefore, the memory requirement and computational cost are significantly reduced in sparse convolutional networks. 
To further reduce the complexity during inference, network binarization can be leveraged for weight and activation quantization. In a 1-bit sparse convolutional layer, both convolutional kernels and activations are binarized to $-1$ and $+1$. In this way, the time-consuming floating-point matrix multiplication can be replaced by bitwise XNOR and popcount operations:
\vspace{-1mm}
\begin{equation}
    \boldsymbol{A^l_b}=sign({\rm popcount}({\rm XNOR}(\boldsymbol{W^l_b}, \boldsymbol{A^{l-1}_b})))
    \vspace{-1mm}
\end{equation} where $\boldsymbol{A^l_b}$ and $\boldsymbol{W^l_b}$ represent the binarized activations and weights in the $l_{th}$ layer respectively, and $\boldsymbol{W^l_b}$ is defined as the binarzed version of the real-valued latent weights $\boldsymbol{W^l_r}$ via $\boldsymbol{W^l_b}=sign(\boldsymbol{W^l_r})$.

\subsection{Shifted Sparse Convolution}\label{a2}

\begin{figure}[t]
    \centering
    \vspace{-2mm}
    \includegraphics[width=1.0\linewidth]{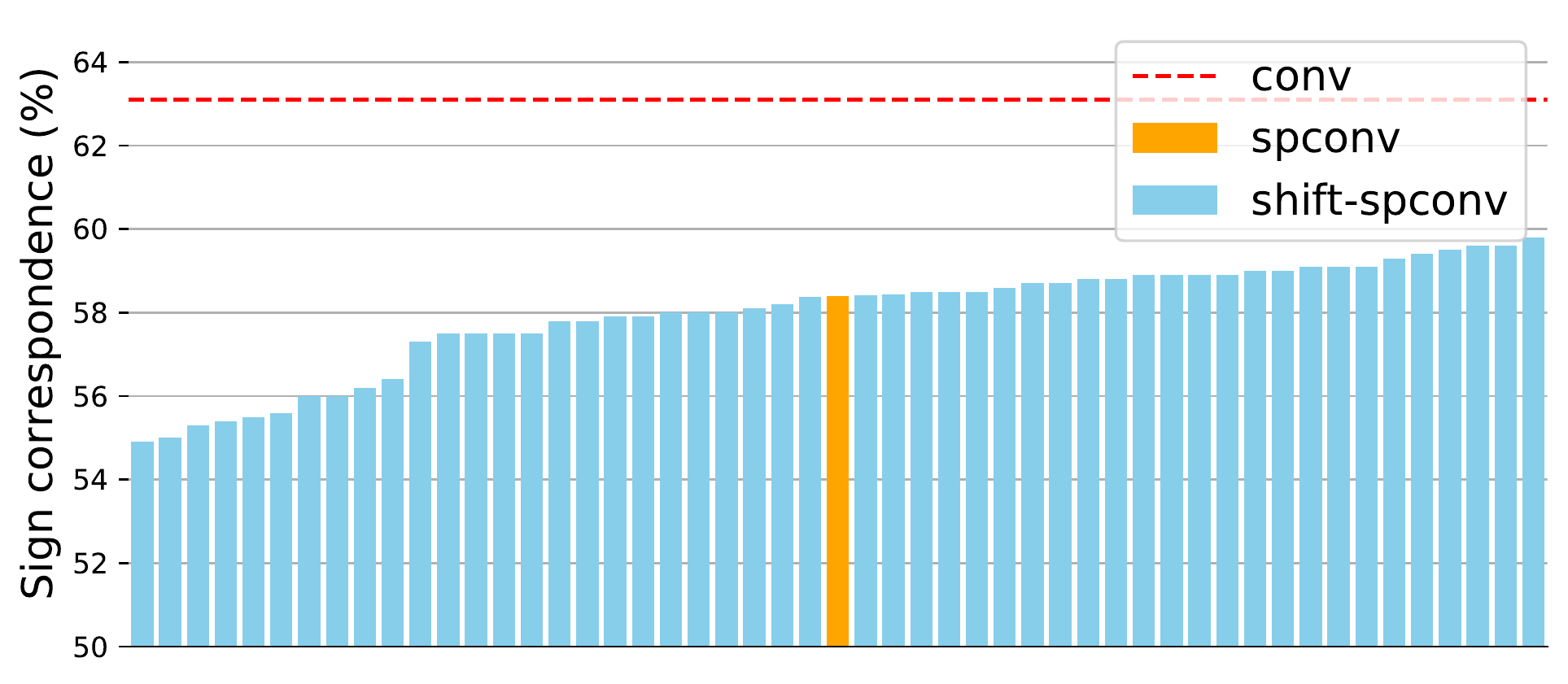}
    \vspace{-6mm}
    \caption{Sign correspondence of activations for the first binary layer when binarizing convolutional network, sparse convolutional network and shifted sparse convolutional network for point cloud segmentation on ScanNet dataset. All networks share the same kernel weights. We sort x-axis (different patterns of sparse convolution) by their sign correspondence for better visualization.}
    \vspace{-2mm}
    \label{fig:toy}
\end{figure}

Since the fixed operation in sparse convolution is only activated when the central input in the receptive field is active, the constrained exploration of the neighbor active sites makes sparse convolutional networks less robust to binarization.
To show this, we calculate the \emph{sign correspondence} (the proportion of activations in binary network that own same signs with the corresponding real-valued activations, which can measure the quantization error as proved in \cite{rastegari2016xnor}) for convolutional network and sparse convolutional network with inputs from the ScanNet dataset. We choose the activations of the first binary layer to avoid the accumulation of quantization errors and adopt the same kernel weights for both networks. As shown in Figure \ref{fig:toy}, the sign correspondences for convolutional layer and sparse convolutional layer are 63.1\% and 58.4\% respectively, which confirms that sparse convolution will bring larger quantization errors than standard convolution.

However, it is infeasible to adopt convolutional layers in point cloud analysis networks for reducing quantization errors due to the large computational cost from growing active sites. As an alternative, we try to explore the subset of convolution. For a single active site, a $3\times3\times3$ convolution kernel will operate 27 times while sparse convolution kernel only operates at the center. What if we keep the same number of operations with sparse convolution but operates at other location? To answer it, we extend sparse convolution to enable it to active at different locations.
Here we propose the shifted sparse convolution(SFSC) shown in Figure \ref{fig:SSSC}(b), which is defined as: 
\vspace{-2mm}
\begin{equation}
    F_k(\boldsymbol W,{\bf x_u})=\sum_{{\bf i}\in N^D({\bf u}+s_k)}{\boldsymbol W_{\bf i}{\bf x_{u+i}}}
\end{equation}
\vspace{-4mm}
\begin{equation*}
    s_k\in\mathbb R^3,\ k\in\{1,2,...,n_s\}
    \vspace{-1mm}
\end{equation*}
where ${\bf u}+s_k$ is the center of shifted cube instead of ${\bf u}$. 
$N^D({\bf u}+s_k)$ is then comprised of the offsets in the shifted cube w.r.t. $\bf u$. 
$n_s$ is the number of all unique shifts. 
For example, for a $3\times3\times3$ sparse convolution operation, there are up to $3^3-1=26$ possible shifts. 

For a general sparse convolution operation, it conducts convolution only when the kernel center overlaps with active sites. 
While in our SFSC operation, the kernel center can shift to any other locations of the kernel. 
We use $\boldsymbol F_{n_s}=\{F_0, F_1, F_2, ..., F_{n_s}\}$ to represent the set of all SFSC operations. Note that we consider the general sparse convolution as a special case of SFSC ($F_0$).  
In a SFSC layer, instead of applying the same sparse convolution operation for all output channels as in a general sparse convolutional layer, we uniformly divide the output channels into several groups (namely channel group), each with a specific SFSC operation. It can be formulated as:
\vspace{-1mm}
\begin{equation}\label{e4}
    y={\rm concat}(f_1(\boldsymbol W_1,x), ..., f_{n_g}(\boldsymbol W_{n_g},x)),\ f_i\in \boldsymbol F_{n_s}
    \vspace{-1mm}
\end{equation}
where $x$ and $y$ are the input and output of this layer. $n_g$ indicates the number of channel groups. $W_i$ refers to the weights for the i-th SFSC operation. The outputs of all SFSC operations are concatenated along the channel dimension, resulting in a tensor with the same shape as the output of a general sparse convolutional layer.

We randomly sample 50 shift configurations for SFSC layers and compute the sign correspondence, which is shown in Figure \ref{fig:toy}. It can be seen that different SFSC layers vary a lot in quantization errors and a proportion of them are more robust to binarization compared to sparse convolutional layer. In another word, if we can find out the (near) optimal configurations for all SFSC layers in a network, the quantization error can be reduced without additional computational cost.

\subsection{Efficient Search for Shift Operation}\label{a3}
Due to the huge design space of shift operation, it is infeasible to decide an optimal configuration for the whole network: the shifted channels and shift directions may be different in each layer, and the total number of possible architectures will be $(8^4)^{13}=9.1\times10^{46}$ for a network with 13 SFSC layers, each layer with 4 channel groups and 8 available shift directions. Although manually designed BSC-Net, which shares the same shift strategy in all SFSC layers, is able to reduce the impact of binarization on the network performance, we resort to automatic architecture search for a better performance. In this section, without further explanation, the default kernel size for original sparse convolution and SFSC is $3\times3\times3$.

\begin{figure}[t]
    \centering
    \vspace{-2mm}
    \includegraphics[width=1.0\linewidth]{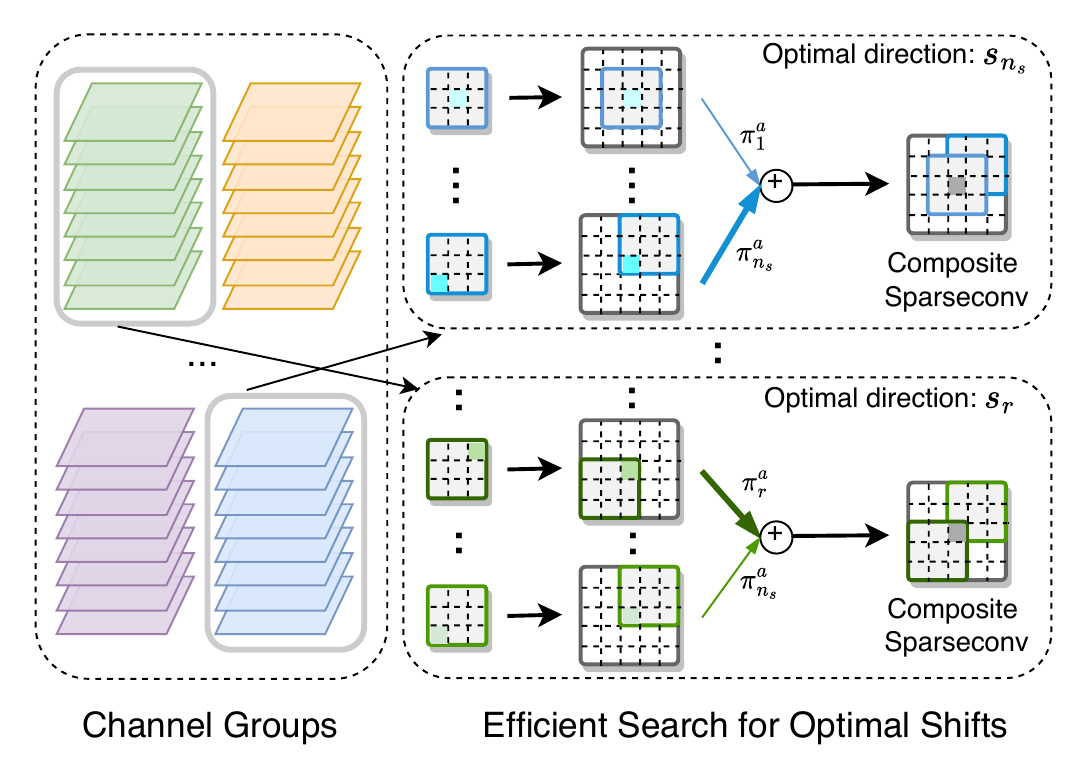}
    \vspace{-6mm}
    \caption{Demonstration of our efficient search method for shift operation. For each SFSC layer and each channel group, we combine all the shift operations in the search space into a $5\times5\times5$ sparse convolution and assign each direction with a soft selector indicating the importance of the corresponding shift operation, which enables us to directly search the best shift operations via end-to-end gradient descent. $\oplus$ stand for summation.}
    \vspace{-2mm}
    \label{fig:SSSC_NAS}
\end{figure}

In our BSC-Net, the optimal shift direction for each channel group and each layer may differ. Thus the problem is to search the optimal shift direction for each channel group in the SFSC layer. We formulate this by searching the optimal $f_i$ in (\ref{e4}):
\vspace{-2mm}
\begin{equation}\label{eq3}
    f_i=\sum_{j=1}^{n_s}o_{ij}^aF_j,\ i\in\{1,2,...,n_g\}
\end{equation}
\vspace{-4mm}
\begin{equation*}
    {\rm s.t.}\ \sum_j{o_{ij}^a}=1,\ o^a\in\{0,1\}.
    \vspace{-2mm}
\end{equation*}
where $o^a$ is a binary selector of the shift direction.
As searching in a discrete space makes it hard to optimize the choices, we reformulate the discrete search space as a continuous one by switching $f_i$ to a composite function $f_i^*$:
\vspace{-2mm}
\begin{equation}\label{eq4}
    f_i^*=\sum_{j=1}^{n_s}\pi_{ij}^aF_j,\ i\in\{1,2,...,n_g\}
\end{equation}
\vspace{-4mm}
\begin{equation*}
    {\rm s.t.}\ \pi^a\in[0,1],\ \pi^a_{ij}=\frac{1}{1+{\rm exp}(-\alpha_{ij})}
    \vspace{-1mm}
\end{equation*}
where the constraints on weight $\pi^a$ are eliminated by introducing a set of real architecture parameters $\{\alpha_{ij}\}$. This sigmoid relaxation~\cite{chu2020fair} will not introduce competition among different SFSC operations as in softmax relaxation~\cite{cai2020rethinking}, which we find to be a better way to search for BSC-Net.
In this way, the composition of SFSC operations are learned by gradient descent in the space of continuous real parameters $\{\alpha_{ij}\}$, which can be optimized end-to-end efficiently.

However, according to (\ref{eq4}), the computation and memory increase linearly with the size of search space. All available SFSC operations need to be conducted in weighted summation $f_i^*=\sum_{j=1}^{n_s}\pi_{ij}^aF_j$. Moreover, each SFSC layer owns different parameters, increasing the difficulty of network optimization. To this end, we propose an efficient search method, which absorb all the operations in search space into a larger sparse convolution, as shown in Figure \ref{fig:SSSC_NAS}.

In this way, we convert the SFSC layer into a $5\times5\times5$ composite sparse convolutional layer, which is used to construct a supernet. This enables us to efficiently search the optimal architecture parameters by end-to-end optimization, regardless of the search space. However, it should be clarified that although the size of search space will not affect the computational efficiency of the supernet, a large search space will make the optimization of architecture parameters hard to converge, thus deteriorate the final performance.

Once the supernet is converged, the optimal BSC-Net must be derived by discretizing the soft selector variables $\pi^a$ of (\ref{eq4}) into the binary selectors $o^a$ required by (\ref{eq3}). 
In order to make sure the performance of supernet can precisely reflect the capability of BSC-Net, we constrain $\pi^a$ in each SFSC layer by a confidence loss:
\vspace{-2mm}
\begin{equation}\label{eq_c}
    L_c=-\frac{1}{n_g \cdot n_s}\sum_i^{n_g}{\sum_j^{n_s}{|\pi_{ij}-0.5|}}
    \vspace{-2mm}
\end{equation}
which pushes $\pi^a$ to discrete values.

\textbf{Optimization approach:}
In order to decouple the weights and architecture parameters for robust learning \cite{cai2020rethinking}, we adopt an alternating optimization approach: 1) fix the $\{\alpha_{ij}\}$ and optimize $\{\boldsymbol W_i\}$; 2) fix $\{\boldsymbol W_i\}$ and update $\{\alpha_{ij}\}$. 

When we derive the BSC-Net from a converged supernet, both weights and architecture parameters need to be considered. Here we find the following strategy works best: we first train the supernet with binary weight and activation to search for the optimal architecture parameters, from which we choose the shift directions with the highest architecture parameters. Then we initialize the searched BSC-Net with the weights from the supernet and follow the same training procedure as our baseline (introduced in Section \ref{exp}).

\section{Experiment}\label{exp}
To investigate the performance of the proposed method, we conduct experiments on several indoor scene datasets including NYU Depth v2 (NYUDv2) \cite{Silberman2012nyu} and ScanNet \cite{dai2017scannet}. We first introduce the datasets, evaluation metrics and implementation details, which is followed by a strong baseline designed for binarization of sparse convolution networks. Then we compare our BSC-Net with the state-of-the-art network binarization methods on sparse convolutional networks. Finally we design ablation studies to show the effectiveness and efficiency of the presented BSC-Net.

\begin{table*}[t]
  \centering
  \small
  \caption{The mIoU of binarzed sparse convolutional networks on ScanNet of different baseline techniques, where the UNET architectures are leveraged. The methods from left to right indicate (1) removing all the skip connections; (2) replacing PReLU with Hardtanh; (3) calculating scaling factor for both activations and weights; (4) using STE to approximate the gradient; (5) removing the skip connections for downsampling/upsampling layers; (6) directly training network with binary weights and activations.}\label{tbl:baseline}
  \begin{tabular}{l|c|c|c|c|c|c|c}
  \noalign{\smallskip}
  \hline
  \hline
  Method & Simplify BS & Simplify AF & Modify SF & Simplify GA & Simplify DS/US & Simplify Init. &Full baseline  \\
  \hline
  mIoU (\%) &37.4 &50.5 &46.1 &49.9 & 47.3/48.7 &34.1 &\textbf{51.7}  \\
  \hline
  \hline
  \noalign{\smallskip}
  \noalign{\smallskip}
  \end{tabular}
  \vspace{-0.5cm}
\end{table*}

\subsection{Experimental Settings}\label{e1}
\textbf{Datasets and metrics:} We conduct experiments on two indoor datasets including NYU Depth v2 (NYUDv2) \cite{Silberman2012nyu} and ScanNet \cite{dai2017scannet}. NYUDv2 contains 1,449 RGB-D scene images, where 795 images are split for training and 654 images for testing. Following \cite{graham20183d}, we adopt 40-class setting where all pixels are labeled with 40 classes and convert the RGB-D images into 3D point clouds. As the horizontal and vertical directions of spatial dimensions in the RGB-D images are discrete, we voxelize the 3D point clouds to 1cm bins by only discretizing the depth dimension. ScanNet consists of 1513 reconstructed indoor scenes with 21 categories, which are split into 1201 and 312 scenes for training and validation respectively. We adopt two popular settings of ScanNet containing \emph{2cm} voxelization and \emph{5cm} voxelization as done in \cite{choy20194d}.

We report the mean intersection of union (mIoU), mean per-point classification accuracy (mAcc) and overall point-wise classification accuracy (Acc) for NYUDv2. For ScanNet, we report mIoU and mAcc. We use the same calculation method in \cite{liu2020react} to count the binary operations (BOPs) and floating point operations (FLOPs), where the total operations for model computation complexity evaluation is counted by OPs = BOPs/64 + FLOPs. The storage cost are measured by summing the number of real-valued parameters and that of binary numbers divided by $32$.

\noindent\textbf{Implementation details:} Following \cite{graham20183d}, we adopt different network architectures for the various datasets. For NYUDv2, we perform experiments with FCN \cite{long2015fully} networks in different sizes, namely FCN-S (small) and FCN-H (huge). 
For ScanNet, we leverage the U-Net \cite{ronneberger2015u} architecture in small and huge sizes represented as UNET-S and UNET-H. The small and huge models differ in numbers of filters and sparse convolutional layers per level, which results in capacity variations of point cloud analysis.
We use Adam with a stepwise scheduler to optimize the network parameters.
The training hyperparamters are introduced in the supplementary materials in detail. We perform data augmentation by applying random affine transformers to the point cloud.

For our BSC-Net, the shift distance in SFSC operations is set to one and the number of channel groups which employ different shift directions is assigned to $8$. The search space of directions contains shifting to $8$ operations represented by $(\pm1, \pm1, \pm1)$ and staying still without shift. Limiting the search space of shift directions for channel groups in each layer significantly reduces the search difficulty while maintains the exploration capability. We also evaluated two variations of our BSC-Net called BSC-Baseline and BSC-Manual to demonstrate the effectiveness of the presented techniques. BSC-Baseline represents the framework that binarizing the sparse convolutional networks with all beneficial recently advances combined (refer to Section \ref{e2}), and BSC-Manual stands for the network binarization for network consisted of SFSC layers with manually defined shift configurations instead of the searched ones. In BSC-Manual, We set $\frac{1}{2}$ of the channel groups unshifted, $\frac{1}{4}$ shift to $(+1, +1, 0)$ and $\frac{1}{4}$ shift to $(-1, -1, 0)$ for NYUDv2, and $\frac{1}{2}$ of the channel groups unshifted, $\frac{1}{4}$ shift to $(+1, +1, +1)$ and $\frac{1}{4}$ shift to $(-1, -1, -1)$ for ScanNet. BSC-Baseline and BSC-Manual are trained in the same way, while BSC-Net is trained with an additional searching stage first.

\subsection{Strong Baseline}\label{e2}
Since network binarization degrades the performance sizably, techniques for accuracy improvements have been studied in recent works of model quantization. To show the performance improvement comes from our proposed method rather than other tricks, we build a strong baseline for binarizing sparse convolutional networks from the recently advances.
Through the empirical study shown in Table \ref{tbl:baseline}, we discover the beneficial techniques for performance enhancement and list as below.


\noindent\textbf{Block sturcture: }We use the same block structure as ReActNet \cite{liu2020react}, where the operations are ordered as Binarization$\rightarrow$SparseConv$\rightarrow$BatchNorm$\rightarrow$Activation in each basic block.

\noindent\textbf{Activation function: }PReLU \cite{he2016deep,liu2020react} considers the negative inputs with better convergence, and we substitute all ReLU activation layers with PReLU to strengthen the performance.

\noindent\textbf{Scaling factor: }We only calculate the layer-wise scaling factor for weights as demonstrated in \cite{martinez2020training}, which is the mean absolute value offull-precision weights.

\noindent\textbf{Gradient approximation: }A piecewise polynomial function \cite{liu2018bi} is used to approximate the sign function, which acquires more accurate gradient during back propagation.

\noindent\textbf{Downsampling/upsampling: }Following \cite{liu2018bi}, the skip connection for downsampling layer is composed of an average pooling land a real-valued convolutional layer with kernel size $1$. We also verify that an unpooling layer with a full-precision convolution with kernel size $1$ is beneficial in the skip connection for upsampling layer.

\noindent\textbf{Initialization: }We first pretrain the network with full-precision weights and activations for initialization. Then the model with binary weights and activations is trained for binarization.

\begin{table*}[t]
  \setlength{\tabcolsep}{6pt}
  \centering
  \small
  \caption{Semantic segmentation results (\%), model storage cost (M) and computation cost in OPs of different methods on ScanNet validation set. Param.\ means the model storage cost (M). \emph{5cm voxel} and \emph{2cm voxel} refer to different resolutions of the input point cloud after voxelization.}
    \begin{tabular}{c|l|c||c|cc||c|cc}  
      \toprule
      & \multirow{2}{*}{Method} & \multirow{2}{*}{Param.} & \multicolumn{3}{c||}{\textbf{5cm voxel}} & \multicolumn{3}{c}{\textbf{2cm voxel}} \\
      & & & OPs & mIoU & mAcc & OPs & mIoU & mAcc \\ 
      \midrule
      \rowcolor{mygray}
    \cellcolor{white} \multirow{9}{*}{\rotatebox[origin=c]{90}{UNET-S}} & Real valued &4.335 &$1.21\times10^9$ & 65.2 & 73.3 &$5.32\times10^9$ & 68.7 & 78.5 \\
      & XNOR-Net &0.136 &$8.07\times10^7$ &33.3 &38.9 &$3.79\times10^8$ &21.0 &26.1  \\
      & XNOR-Net++ &0.136 &$8.07\times10^7$ &12.6 &15.9 &$3.79\times10^8$ &11.2 &13.7  \\
      & BiPointNet &0.136 &$8.07\times10^7$ &30.1 &36.2 &$3.79\times10^8$ &18.4 &20.7  \\
      & Bi-Real-Net &0.138 &$8.12\times10^7$ &48.3 &56.6 &$3.82\times10^8$ &51.2 &63.3 \\
      & ReActNet &0.138 &$8.12\times10^7$ &43.6 &50.2 &$3.82\times10^8$ &46.9 &52.9 \\
      & BSC-Baseline &0.139 &$8.12\times10^7$ &51.7 &61.8 &$3.82\times10^8$ &54.9 &65.3 \\
      & BSC-Manual &0.139 &$8.12\times10^7$ &53.2 &63.7 &$3.82\times10^8$ &57.8 &66.6 \\
      & BSC-Net &0.139 &$8.12\times10^7$ &54.4 &65.2 &$3.82\times10^8$ &61.4 &70.4 \\
      \midrule
      \rowcolor{mygray}
    \cellcolor{white} \multirow{9}{*}{\rotatebox[origin=c]{90}{UNET-H}} & Real valued &30.104 &$7.65\times10^{9}$ & 67.6 & 75.1 &$3.38\times10^{10}$ & 71.0 & 79.0 \\
      & XNOR-Net &0.939 &$3.61\times10^{8}$ &46.6 &53.5 &$1.75\times10^{9}$ &34.9 &40.1  \\
      & XNOR-Net++ &0.939 &$3.61\times10^{8}$ &13.3 &16.4 &$1.75\times10^{9}$ &12.8 &16.2 \\
      & BiPointNet &0.939 &$3.61\times10^{8}$ &45.2 &52.4 &$1.75\times10^{9}$ &34.3 &39.8 \\
      & Bi-Real-Net &0.948 &$3.63\times10^{8}$ &53.4 &63.2 &$1.76\times10^{9}$ &57.3 &66.9 \\
      & ReActNet &0.949 &$3.63\times10^{8}$ &52.2 &59.0 &$1.76\times10^{9}$ &57.1 &67.0 \\
      & BSC-Baseline &0.952 &$3.63\times10^{8}$ &56.0 &65.9 &$1.76\times10^{9}$ &59.3 &68.3 \\
      & BSC-Manual &0.952 &$3.63\times10^{8}$ &59.3 &68.7 &$1.76\times10^{9}$ &62.2 &70.1 \\
      & BSC-Net &0.952 &$3.63\times10^{8}$ &62.2 &70.5 &$1.76\times10^{9}$ &63.9 &71.6 \\
      \bottomrule
    \end{tabular}
  \vspace{-2.0mm}
  \label{tbl:scannet}
\end{table*}

\begin{table}[t]
  \setlength{\tabcolsep}{3pt}
  \centering
  \small
  \caption{Semantic segmentation results (\%), model storage cost (M) and computation cost in OPs of different methods on NYUDv2 test set. Param.\ means the model storage cost (M).}
    \begin{tabular}{c|l|c||c|ccc}  
      \toprule
      & Method & Param. & OPs & mIoU & mAcc & Acc  \\
      \midrule
      \rowcolor{mygray}
    \cellcolor{white} \multirow{9}{*}{\rotatebox[origin=c]{90}{FCN-S}} & Real valued &2.496 &$1.24\times10^9$ & 33.9 & 47.7 & 64.7 \\
      & XNOR-Net &0.108 &$1.72\times10^8$ &22.1 &32.7 &57.3   \\
      & XNOR-Net++ &0.108 &$1.72\times10^8$ &8.5 &13.5 &43.9  \\
      & BiPointNet &0.108 &$1.72\times10^8$ &24.9 &35.7 &59.3  \\
      & Bi-Real-Net &0.110 &$1.75\times10^8$ &27.3 &38.4 &60.0 \\
      & ReActNet &0.110 &$1.75\times10^8$ &25.4 &36.6 &58.9 \\
      & BSC-Baseline &0.110 &$1.75\times10^8$ &27.8 &39.9 &60.1 \\
      & BSC-Manual &0.110 &$1.75\times10^8$ &28.7 &40.9 &60.2   \\
      & BSC-Net &0.110 &$1.75\times10^8$ &29.7 &42.1 &61.2 \\
      \midrule
      \rowcolor{mygray}
    \cellcolor{white} \multirow{9}{*}{\rotatebox[origin=c]{90}{FCN-H}} & Real valued &10.025 &$4.82\times10^9$ & 36.9 & 50.4 & 67.2 \\
      & XNOR-Net &0.357 &$3.08\times10^8$ &27.1 &38.8 &59.6 \\
      & XNOR-Net++ &0.357 &$3.08\times10^8$ &8.4 &13.6 &43.2 \\
      & BiPointNet &0.357 &$3.08\times10^8$ &28.1 &40.8 &60.1 \\
      & Bi-Real-Net &0.361 &$3.09\times10^8$ &30.4 &41.8 &61.1  \\
      & ReActNet &0.361 &$3.09\times10^8$ &27.0 &40.1 &58.9 \\
      & BSC-Baseline &0.362 &$3.09\times10^8$ &32.0 &44.4 &63.3 \\
      & BSC-Manual &0.362 &$3.09\times10^8$ &32.5 &45.1 &63.5  \\
      & BSC-Net &0.362 &$3.09\times10^8$ &33.9 &46.2 &64.5 \\
      \bottomrule
    \end{tabular}
  \vspace{-2.0mm}
  \label{tbl:nyu}
\end{table}

\subsection{Comparison with State-of-the-art}\label{e3}
In this section, we compare our method with state-of-the-art binarization methods, including XNOR-Net \cite{rastegari2016xnor}, XNOR-Net++ \cite{bulat2019xnor}, BiPointNet \cite{qin2020bipointnet}, Bi-Real-Net \cite{liu2018bi} and ReActNet \cite{liu2020react}. We also provide the performance of the real valued models for reference. Experiments are conducted on NYUDv2 and ScanNet.

\textbf{Results on NYUDv2:}
Table \ref{tbl:nyu} illustrates the comparison of storage, operations per second (OPs) and semantic segmentation results across several popular network binarization methods and our BSC-Net. Bi-Real-Net performs best among previous methods, which shows the gradient approximation method and the skip connection structure are general and effective in sparse convolutional network binarization. Although BiPointNet is designed for 3D point cloud analysis, it fails to achieve satisfactory performance because the operations such as maxpooling and point-wise MLP used in PointNet are not adopted in sparse convolutional networks. BSC-Baseline outperforms the previous methods by a large margin and its performance is further boosted by the proposed SFSC module, i.e.\ BSC-Manual. When adopting the efficient differentiable search method, BSC-Net achieves the state-of-the-art performance in both architectures of FCN, while the extra computataional overhead is negligible compared to previous methods.Observing the last low in Table \ref{tbl:nyu}, The performance gap between real valued FCN-H and our BSC-Net has even been narrowed to less than 3\%, which shows the great application potentials of our method.

\textbf{Results on ScanNet:}
Different from NYUDv2 in which the point clouds are generated from single RGB-D images, ScanNet provides larger and more complete point cloud scenes via 3D reconstruction. Therefore, we can evaluate our BSC-Net on ScanNet with different resolutions of the input point cloud after voxelization as shown in Table \ref{tbl:scannet}. 
Following \cite{graham20183d}, we evaluate all results three times to address the problem of the number of voxels being greatly smaller than that of points.
Similar to NYUDv2, BSC-Baseline outperforms the previous state-of-the-art. We found the gap between BSC-Baseline and previous methods were larger because the upsampling layer in UNET is implemented by deconvolution, which is more sensitive to binarization than the interpolation used in FCN. In each setting, BSC-Manual gains consistent improvement over BSC-Baseline, and BSC-Net further achieves state-of-the-art performances which proves the effectiveness of differentiable search strategy.
We also noticed that the improvement of BSC-Net over BSC-Baseline on ScanNet is larger than that on NYUDv2, that shows our method can exploit richer geometric information from 3D point clouds than 2.5D depth map.

\subsection{Ablation Study}\label{e4}
We conduct ablation studies to show how different hyperparamters and strategies influence the performance of the proposed BSC-Net. We study the effects of the search space and number of channel group as well as searching strategy in our differentiable search method on the final performance. The experiments are conducted on ScanNet (\emph{5cm} voxelization) using UNET-S.



\textbf{Performance w.r.t. searching hyperparamters:}
In order to reduce the search cost as well as the optimization difficulties, we search the optimal shift strategies for different layers in BSC-Net from a subset of the whole search space, and we also partition the channels into groups which share the same shift strategies. Table \ref{spgn} demonstrates the performance variation for BSC-Net with different search space and group numbers of SFSC operations, where $S$ and $C_k^d$ represent the convolutions staying still and those shifted to the direction of the k-th vertex of a cube.

\begin{table}[t]
  \centering
  \setlength{\tabcolsep}{4pt}
  \caption{The effects of search space and group number in differentiable search method on the final performance.}\label{spgn}
  \vspace{-2mm}
  \small
  \begin{tabular}{lccc}
  \noalign{\smallskip}
  \hline
  \hline
  Search space &Group number \ \  &mIoU(\%) &mAcc(\%)  \\
  \hline
  Baseline: $\{S\}$ &-- &51.7 &61.8 \\
  \hline
  $\{S,C_1,C_8\}$ &8 &53.6 &63.9  \\
  $\{S,C_1,C_4,C_6,C_7\}$ &8 &54.2 &64.9  \\
  \hline
  \multirow{3}*{$\{S,C_1,C_2,C_3,$} &2 &52.9 &63.8  \\
  &4 &53.4 &64.6 \\
  \multirow{2}*{$C_4,C_5,C_6,C_7,C_8\}$}&8 &\textbf{54.4} &\textbf{65.2} \\
  &16 &54.0 &64.5 \\
  \hline
  \hline
  \noalign{\smallskip}
  \noalign{\smallskip}
  \end{tabular}
  \vspace{-2mm}
\end{table}

\begin{table}[t]
  \centering
  \setlength{\tabcolsep}{0.6pt}
  \caption{The effects of relaxation and derivation strategy in differentiable search method on the final performance. $^*$ indicates the architecture parameters are frozen.}\label{srtg}
  \vspace{-2mm}
  \small
  \begin{tabular}{cccc}
  \noalign{\smallskip}
  \hline
  \hline
  Relaxation &Derivation \ \  &mIoU(\%) &mAcc(\%)  \\
  \hline
  Random &$D(32,32)\rightarrow D(1,1)$ &51.5 &61.2 \\
  \hline
  \multirow{3}*{Softmax} &$S(32,32)\rightarrow S(1,1)\rightarrow D(1,1)$ &51.8 &62.0 \\
  &$S^*(32,32)\rightarrow S(1,1)\rightarrow D(1,1)$ &52.3 &63.2 \\
  &$S(1,1)\rightarrow D(32,32)\rightarrow D(1,1)$ &53.5 &64.5 \\
  \hline
  \multirow{3}*{Sigmoid} &$S(32,32)\rightarrow S(1,1)\rightarrow D(1,1)$ &52.6 &64.0 \\
  &$S^*(32,32)\rightarrow S(1,1)\rightarrow D(1,1)$ &53.7 &64.7 \\
  &$S(1,1)\rightarrow D(32,32)\rightarrow D(1,1)$ &\textbf{54.4} &\textbf{65.2} \\
  \hline
  \hline
  \noalign{\smallskip}
  \noalign{\smallskip}
  \end{tabular}
  \vspace{-2mm}
\end{table}

Observing the third, fourth and seventh rows, we conclude that the mIoU and mAcc of BSC-Net improves as the size of the search space increases. The improvement from search space in size 3 to that in size 5 is much higher than that from size 5 to size 9, which indicates that large search space causes optimization difficulties in differentiable search method and a subset of the whole search space contains the solution near to the optimal one. According to the last four rows in Table \ref{spgn}, the performance achieves the optimal for medium numbers of groups in the differentiable search. Increasing the numbers causes the optimization difficulties due to the large search space, and decreasing the numbers excludes the promising solution due to the channel correlation.

\textbf{Performance w.r.t. searching strategy:}
We further investigate the effects of searching strategy in Table \ref{srtg}, including relaxation method for the binary selector in (\ref{eq3}) and strategy for deriving BSC-Net from the supernet. For softmax relaxation, $\{\pi^\alpha_{ij}\}$ are defined as $\pi^a_{ij}=\frac{{\rm exp}(\alpha_{ij})}{\sum_k{{\rm exp}(\alpha_{ik})}}$, and the confidence constraint in (\ref{eq_c}) is changed to:
\vspace{-2mm}
\begin{equation}
  L_c=-\sum_i^{n_g}{\sum_j^{n_s}{I_{ij}log\pi^a_{ij}}},\ I_{ij}=\begin{cases}
      1, & j=\mathop{\text{argmax}}\limits_{k}\ \pi^a_{ik} \\
      0, & \text{otherwise}
  \end{cases}
  \vspace{-1mm}
\end{equation}
which pushes $\pi^a$ to a one-hot tensor for better derivation. We use $S(W,A)$ and $D(W,A)$ to represent supernet and derived BSC-Net with W-bit weights and A-bit activations.

We first conduct random assignment on the shift directions for each layer and channel group. As shown in the first row, the performance of BSC-Net falls even behind of the baseline, which shows searching or manually designing a proper SFSC configuration is essential for realizing the potential of BSC-Net. The last six rows show that sigmoid relaxation is better than softmax, which is due to softmax will bring competition between different SFSC operations and hurts the performance of supernet. The results also testify the superiority of our derivation strategy.

Notably, we should emphasize that the SFSC operation is proposed for reducing quantization error, which does not work for real-valued networks. We do not observe an obvious change in performance when equipping the real-valued UNET or FCN with SFSC.

\subsection{Visualization Results}\label{e5}
\begin{figure}[t]
\centering
\includegraphics[width=1.0\linewidth]{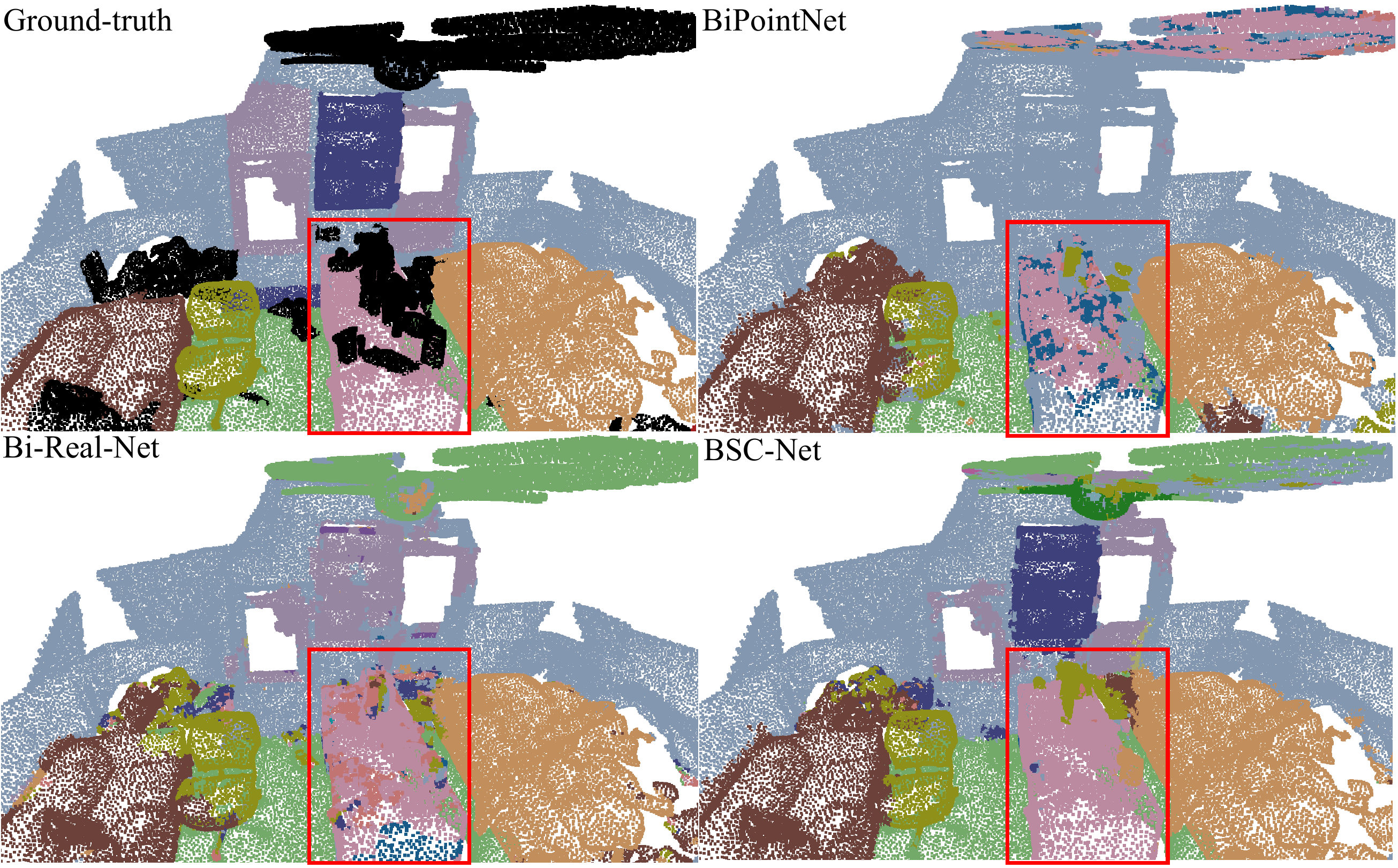}
\vspace{-6mm}
\caption{Visualization results of different methods.}
\label{fig:vis}
\vspace{-4mm}
\end{figure}

We viusalize the segmentation prediction for different methods in Figure \ref{fig:vis}. Black regions in the ground-truth refer to undefined categories. The predictions of previous methods are discontinuous and they misclassify the shelf as wall or window, while our BSC-Net outputs smooth and accurate predictions. The results in red boxes also provide an intuitive explanation on our method: when binarized, sparse convolutional networks fail to fully explore the neighbor active sites with increased quantization errors and thus predict discontinuous results. On the contrary, BSC-Net better exploits the neighborhood and reduces the quantization error.

\section{Conclusion}
In this paper, we have presented BSC-Net that learns binary sparse convolutional networks for efficient point cloud analysis. We present the shifted sparse convolution that is activated for receptive field whose pre-defined locations match active sites. By searching the optimal positions for active site matching in shifted sparse convolution, the quantization errors in binarized sparse convolutional networks are alleviated significantly without additional computational cost. For fair evaluation, we combine previous techniques to construct a strong baseline. Extensive experiments on semantic segmentation of point clouds demonstrate the superiority of BSC-Net.

\section*{Acknowledgements}
This work was supported in part by the National Key Research and Development Program of China under Grant 2017YFA0700802, in part by the National Natural Science Foundation of China under Grant 62125603, and in part by a grant from the Beijing Academy of Artificial Intelligence (BAAI).

\begin{figure*}[t]
    \centering
    \includegraphics[width=1.0\linewidth]{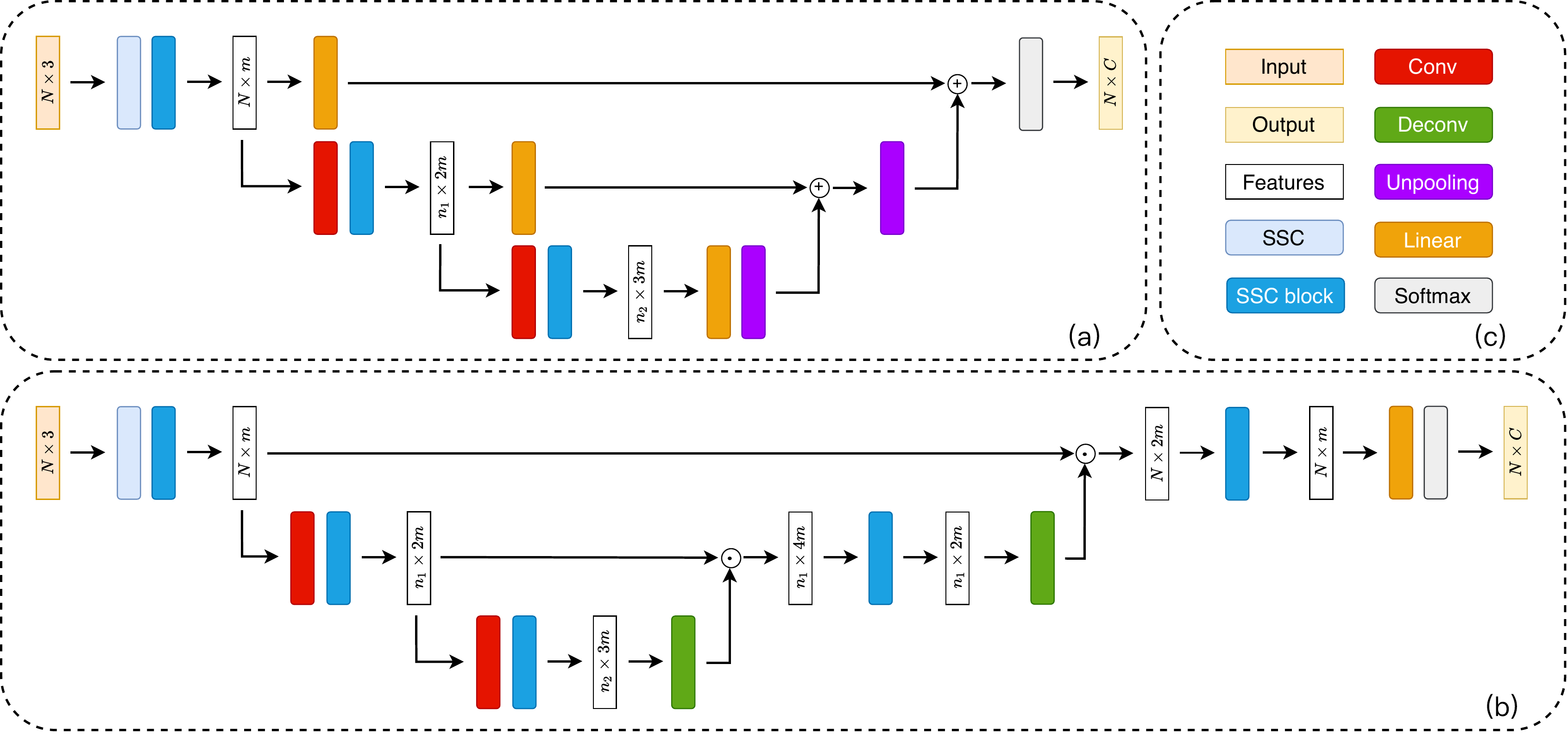}
    \caption{The overall frameworks of FCN (a) and UNET (b) which are constructed with the basic blocks in (c).}
    \label{arch}
\end{figure*}

\appendix
\section*{Supplementary Material}
\section{Overview}
In this supplementary material, we detail the network architectures and training hyperparamters used in our experiments. Section \ref{A} shows how to construct the whole network with basic blocks. Section \ref{B} details the training hyperparamters used in our experiments.

\section{Network Architecture}\label{A}
\subsection{Overall Framework}

We illustrate the architectures of FCN and UNET in Figure \ref{arch}. Only two levels of downsampling/upsampling are shown in the figure for simplicity. Note that for each level only one SSC block is drawn, while in fact there may be one or two blocks according to the size of networks.

FCN-S has 16 filters in the input layer, and one SSC block per level. FCN-H has 24 filters in the input layer, and two SSC blocks per level. Both networks use eight levels of downsampling and upsampling. We increase the number of filters in the networks when downsampling: in particular, we add 16 (S) or 24 (H) filters every time we reduce the scale. 
UNET-S has 16 initial filters and one SSC block per level. UNET-H has 32 initial filters and two SSC blocks per level. Both networks use six levels of downsampling and upsampling. Each downsampling operation adds 16 (S) or 32 (H) filters, while upsampling operation subtracts the same numbers of filters correspondingly.

\begin{figure*}[t]
    \centering
    \includegraphics[width=1.0\linewidth]{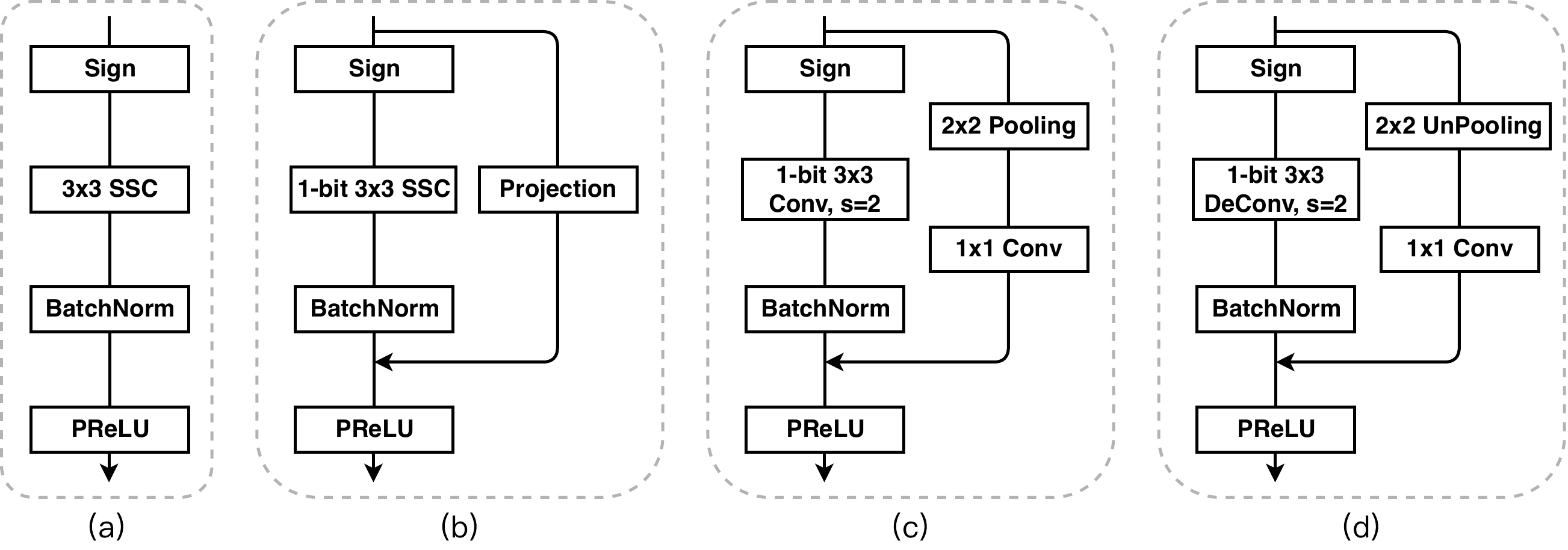}
    \caption{Details of the basic blocks. (a) SSC layer, (b) SSC block, (c) Conv layer, (d) DeConv layer.}
    \label{block}
\end{figure*}

When binarizing the network, the first SSC layer and the linear layers are kept real-valued following previous methods~\cite{liu2018bi,liu2020react,martinez2020training}. We adopt our SFSC module in the SSC block for BSC-Manual and BSC-Net, which is detailed in next subsection.

\subsection{Block Detail}
We detail the structure of basic blocks contained in the binary FCN and UNET in Figure \ref{block}. For SSC layer, the projection indicates identity mapping when the input and output channel are equal, otherwise it is a 1-bit $1\times 1$ convolution operation. For Conv and DeConv blocks, we keep the $1\times 1$ convolution operation real-valued, which is proved to be essential for binarizing performance~\cite{liu2018bi,martinez2020training}.

\section{Training Hyperparamters}\label{B}
There are 2 training stages when training BSC-Baseline and BSC-Manual while 3 stages when training BSC-Net. For all stages, we set max epoch as 128, weight decay as 0 and adopt Adam optimizer with a stepwise scheduler which steps at 60 and 100 epoch (reduce the learning rate by a factor of 10).
The initial learning rates for the first and second stage when training BSC-Baseline and BSC-Manual are set to 0.001 and 0.0002. While for training BSC-Net, the initial learning rates for the three stages are set to 0.001, 0.001 and 0.0002. The weight of the confidence loss is set to 0.1 for FCN and 0.01 for UNET. Note that for training UNET-H on \emph{2cm} voxel,
we double the max epoch and the stepping epochs as the huge network is hard to converge.

{\small
\bibliographystyle{ieee_fullname}
\bibliography{egbib}
}

\end{document}